# An Evaluation of Methods for Real-Time Anomaly Detection using Force Measurements from the Turning Process

Yuanzhi Huang[12], Eamonn Ahearne[3], Szymon Baron[4], Andrew Parnell[56]

1. School of Mathematics and Statistics, University College Dublin, Dublin, Ireland
2. Insight Centre for Data Analytics, Dublin, Ireland
3. School of Mechanical and Materials Engineering, University College Dublin, Dublin, Ireland
4. DePuy Synthes, Cork, Ireland
5. Hamilton Institute, Maynooth University, Kildare, Ireland
6. I-Form Centre for Advanced Manufacturing, Maynooth University, Kildare, Ireland

**Corresponding Author:**

Yuanzhi Huang

Email: Yuanzhi.Huang@ucd.ie

Tel: +353 872261979

ORCID ID: 0000-0002-4332-326X

**Abstract:** We examined the use of three conventional anomaly detection methods and assess their potential for on-line tool wear monitoring. Through efficient data processing and transformation of the algorithm proposed here, in a real-time environment, these methods were tested for fast evaluation of cutting tools on CNC machines. The three-dimensional force data streams we used were extracted from a turning experiment of 21 runs for which a tool was run until it generally satisfied an end-of-life criterion. Our real-time anomaly detection algorithm was scored and optimised according to how precisely it can predict the progressive wear of the tool flank. Most of our tool wear predictions were accurate and reliable as illustrated in our off-line simulation results. Particularly when the multivariate analysis was applied, the algorithm we developed was found to be very robust across different scenarios and against parameter changes. It shall be reasonably easy to apply our approach elsewhere for real-time tool wear analytics.

## 1. Introduction

Modern production of high-precision components applying relevant metal cutting technology relies on using sophisticated computer numerical control (CNC) machine tools. These manufacturing processes are typically regarded as high-value add and achieve micron-order dimensional form accuracy. New technologies for optimisation and automation of production value streams are critical to sustaining profitability in highly-competitive industry sectors such as medical devices, automotive and aerospace. These technologies can be generalised as those which directly improve the process performance (i.e. better tooling) or increase the level of process intelligence. To increase the level of intelligence of cyber-physical production cells requires substantial and real-time knowledge of the key aspects of the process. This allows for real-time automation of production, process planning and management (Altintas 2012; Byrne et al. 2016).

Tool deterioration is defined as a loss of the quality of the cutting edge due to an adverse change of its geometry at the tool-chip contact zone as a result of the cutting process operation (ISO 3685 1993). When the quality of the cutting edge is lost, the tool must be replaced. Tooling represents a significant consumable cost and is one of the main causes of variations in the process outputs. Tool deterioration is particularly problematic when machining materials with a low machinability index (“difficult to cut”). Worn tools cannot generate features meeting the form, finish and accuracy specifications and machining with micro-fractured or excessively worn tool can result in an irreversible sub-surface damage and non-conformance of the component. The ability to precisely detect the tool “end of life” can maximize the utilization of this consumable and reduce the amount of tool-wear related scrap and re-processing. Particularly, timely diagnosis of tool breakage can aid in prevention of unscheduled machine tool downtime. Clearly an automated and precise identification of the tool end of life is of major interest to industry.

Currently, the majority of manufacturing processes employ CNC machines with tool changes determined by historical performance and/or the experience or tacit knowledge of the production operators. The time interval can be, for instance, fixed in a batch production environment. This not only inhibits the possible extent of process automation, but also increases the process variability and associated cost of produced goods. Here, on-line quality process control can be greatly improved by introducing tool wear monitoring based on real-time data to optimise tool life in the manufacturing environment.

This work shall present a novel approach to applying anomaly detection and real-time tool condition monitoring to cutting force measurement data from a turning process. In turning, a single point tool removes a radial depth (“of cut”) of material from a rotating bar by a controlled feed rate (measured in µm / rev.) in a direction parallel to the axis of rotation of the bar. The relative motion thus reduced the diameter of the bar and produces a precision cylindrical surface. The machine, process and experimental set-up are described in detail in McParland et al. (2017).  Here, we focus on the data streams of the measured forces.

The time series data used contains both signal information (i.e. patterns and trends superposed from physical phenomena) related to the parameters programmed on the CNC machine and low-amplitude noise (i.e. random fluctuations in force). We can infer from post-processed data that when the used tool is in a poor health condition, there could be one or more surprising anomalous patterns (i.e. anomalies) to be identified in cutting force data streams, suggesting a developed fault, a condition that exceeds the machine’s specification range (e.g. a force overload), or considerable tool wear. An anomalous pattern can be either a cluster of outlier points or an implausible mean/variance deviation (an aberration) that is developing over time, depending on the pre-defined discrimination rules. As we can see in Fig. 1, for instance, force values become more volatile and unstable towards the end of the tool life.

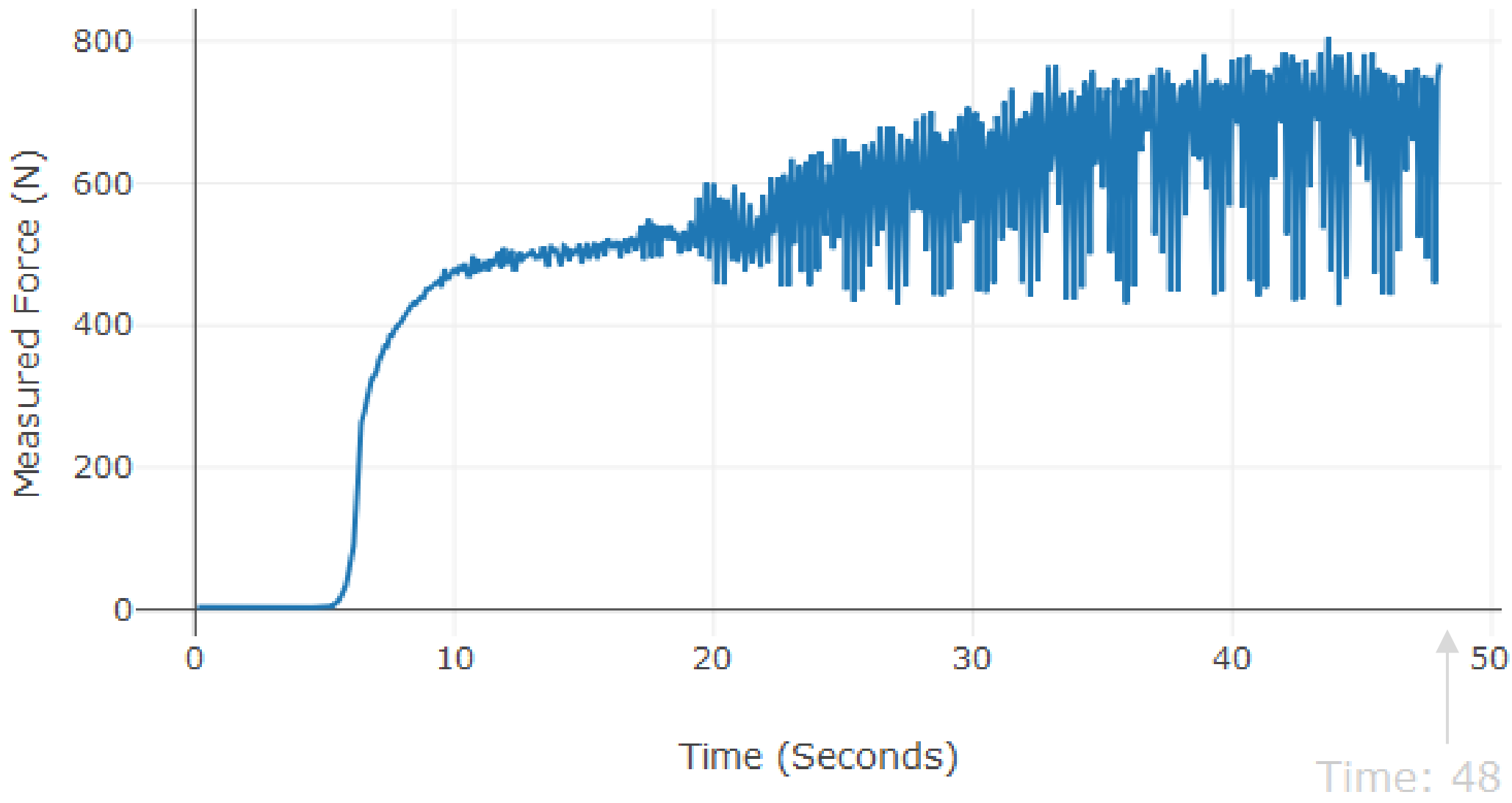

**Fig. 1** A force data stream extracted from the turning process. In this case, we have simulated a real-time data acquisition and visualisation process where anomalies and force spikes start to appear after the first 20 seconds. This would suggest repairing or replacing the tool at the appropriate time (e.g. 30 seconds after the process starts) for improved process control. Otherwise, a damaged tool would affect the machine and lead to substandard components (e.g. with unacceptable levels of surface roughness) in the manufacturing process

To make informed decisions on tool wear management, we use real-time statistical methods to learn from force data streams and recognise potential anomalies in the turning process. We then introduce a tool wear monitoring system that analyses data on a real-time basis and sends warnings whenever tool failure or an unexpected fault is deemed to be imminent. This requires an algorithm capable of fast processing of a volume of streaming data and utilising the extracted information of anomalies to forecast a tool wear threshold (i.e. the point at which the tool wear is going to exceed the tolerance or expected tool life).

The remainder of this paper is organised as follows. In Section 2.1, we discuss how the original data streams are formatted and transferred into our computing environment in R. We examine literature sources for the popular statistical and machine learning methods in Section 2.2, and those feasible for detecting anomalies will be tested in our real-time algorithm discussed in Section 3. The results obtained from our dataset are provided in Section 4, where we also investigate how changing some computer parameters could influence tool wear detection and optimise tool life. Section 5 discusses our results and we then comment on how practitioners could benefit from our approach and our future work to improve the current algorithm framework and work-flow in manufacturing process control. All R code and data are available at GitHub (https://github.com/Yuanzhi-H/Real-Time-Detection-Methods).

## 2. Methods

### 2.1. Data Transfer and Processing

In order to build an informative tool wear monitoring mechanism for machining of cobalt chromium (molybdenum) alloy (Co-Cr-Mo to ASTM F75), we conducted a controlled tool wear experiment on a CNC turning centre instrumented with 3-component force measurement. Potential anomalies should be identified from the continuous force measurement by applying real-time statistical methods, which often indicates significant tool wear. There are similar case studies of industrial damage detection reviewed in Chandola et al. (2009) and Hameed et al. (2009).

In our manufacturing experiment, described in McParland et al. (2017), measurements of the three force components were obtained over the life of a tool in each of a series of test runs which were part of a designed experiment based on a prescribed Sobol sequence where the settings of cutting speed and feed rate were varied in each test run. The force measurements were acquired using a Kistler 50019 charge amplifier, NI DAQ Pad, and LabVIEW software. The force data streams of the turning process can be extracted in real-time from the CNC machine and saved into data files readable in LabVIEW and DIAdem, for instance. These files can be converted into CSV (comma-separated values) format so that we load them into the computing environment of R (R Core Team 2014). While we read all the data for real-time simulations and demonstrations, we test the algorithms off-line as though force data is arriving in streamed packets. This should emulate the real-time environment. Depending on the data volume, a time streaming scheme can be chosen between rolling origin (RO) and rolling window (RW) (Tashman 2000). RO would demand all historical data to be saved for real-time detection during the experiment, whereas RW is more economic and it enables the system to free some space and memory by clearing old data and records regularly. Overall, RO is best for short processes generating limited data and RW is required for longer-term continuous monitoring on one or several machines.

The resultant force of the turning process is made of three components: the tangential force ($F_t$), the thrust force ($F_p$), and the feed force ($F_f$), recorded at a sampling rate of 30kHz from said Kistler piezoelectric dynamometer integrated in the tool support structure for maximum sensitivity to the force flux in three axes. We start by resampling the force measurements at the in-plane natural frequency (3500Hz) of the dynamometer. To efficiently process high-frequency data in the real-time tool wear monitoring environment, the algorithm should quickly build and update a feature vector that contains highly informative summary statistics. As described in Section 3, we choose to calculate the 90% quantiles of data so as to draw a smart 10Hz subsample out of the original data of 3500 Hz, for each run of the design given in Table 1, the order of which follows the Sobol sequence in McParland et al. (2017). As a common practice in machine learning, we divide the 21 runs into three sets as indicated in Table 1: 16 runs for algorithm training; 4 runs for performance tests, and 5 other runs which resulted in milli-scale chipped / fractured cutting edges (hereafter called “gross fracture”).

**Table 1** The 21-Run Experiment: linear spindle speed $v$ (m/min) and feed rate $f$ (µm/rev) are the two factors controlled for the turning process. At the end of each run, four of which are left out as a small test set for validation purpose, the flank wear on the tool was measured (if the wear was progressive), based on which we deduct the time when the flank wear is 150 µm

| Run | Spindle Speed (m/min) | Feed Rate (µm/rev) | VB ≈ 150 µm Time (second) | End of Run Time (second) | Flank Wear (µm) | Test Set |
|---|---|---|---|---|---|---|
| 1 | 40 | 40 | 86.9 | 126 | 223.00 | No |
| 2 | 40 | 30 | 115.9 | 163 | 217.95 | No |
| 3 | 30 | 50 | 625.9 | 879 | 211.24 | No |
| 4 | 35 | 35 | 165.6 | 236 | 215.88 | No |
| 5 | 55 | 55 | 45.0 | 52 | Broken | Broken |
| 6 | 45 | 25 | 107.8 | 141 | 202.01 | Yes |
| 7 | 25 | 45 | 538.3 | 767 | 214.42 | Yes |
| 8 | 28 | 32.5 | 383.4 | 570 | 228.27 | No |
| 9 | 48 | 52.5 | 75.0 | 82 | Broken | Broken |
| 10 | 58 | 22.5 | 48.7 | 63 | 201.53 | No |
| 11 | 38 | 42.5 | 68.4 | 116 | 264.09 | No |
| 12 | 33 | 27.5 | 193.8 | 272 | 212.99 | No |
| 13 | 53 | 47.5 | 20.0 | 42 | Broken | Broken |
| 14 | 43 | 37.5 | 65.7 | 95 | 224.93 | Yes |
| 15 | 23 | 57.5 | 351.6 | 409 | 177.65 | No |
| 16 | 24 | 38.75 | 361.5 | 526 | 221.59 | No |
| 17 | 44 | 58.75 | 30.0 | 50 | Broken | Broken |
| 18 | 54 | 28.75 | 39.5 | 48 | 187.68 | No |
| 19 | 34 | 48.75 | 142.5 | 203 | 220.63 | No |
| 20 | 39 | 23.75 | 144.6 | 190 | 199.62 | Yes |
| 21 | 59 | 43.75 | 55.0 | 159 | Broken | Broken |

As cutting proceeds, and material is removed as a function of time and progressive tool wear increases, we might observe an anomalous feature such as a sudden increase in force. In Table 1, the end-of-run flank wear (VB) exceeds 200 µm most of the time, and the tool flank is grossly fractured in 5 of the 21 cases (Runs 5, 9, 13, 17, and 21). When flank wear is progressive, we make a simplified assumption that it is a function of time (Baron and Ahearne 2017). In real life, however, tool deterioration (in various forms) often accelerates towards the end of tool life while progressive tool wear accelerates also in the initial "run-in" phase. The actual tool failure time is unknown in the turning process, but we assume this to be the time when the flank wear is between 160 and 180 µm, as required in this case. The earliest anomalies should be located somewhere before the failure, when flank wear reaches a certain level and starts to make noticeable impacts on the performance. Here, for the sake of this exercise, we aim to predict the time when $\mathrm{VB} \approx 150\mu m$, i.e. such that the remaining tool life (vs progressive wear) is relatively short, and this can be reset to whatever value we want. Precise identification of this will enable us to further predict tool failure time and be warned in time before failure.

In the case of tool edge gross fracture, as in five of our runs, it is inappropriate to measure flank wear so that reported in Table 1 is the time when anomalies in the data stream are first observed. Because the tool wear mechanism involved differs from that of progressive wear, these runs cannot be included for training. We will however use them for a second test based on comparing our results against the $\mathrm{VB} \approx 150\mu m$ time.

In a real-time environment, one approach is to monitor force components as three independent univariate data streams. An alternative is to integrate these components into a multivariate time series that indicates resultant force. While the input of our real-time monitoring algorithm are the force data streams, the main output will be the time when one or more critical anomalies are found: $\hat{t}_0$ (seconds). To evaluate how accurate our predictions are, we apply a score function based on $\Delta$, the difference in the finished tool life at $\hat{t}_0$ and the assumed $\mathrm{VB} \approx 150\mu m$ time $t_0$ indicated in Table 1. The unit of this cutting distance difference is meter. An industry example of applying a score function is the Numenta Anomaly Benchmark (Lavin and Ahmad 2015). In contrast, our version of the score function is defined on the axis of $\Delta$ as

$$\text{Score} = \begin{cases} \text{MAX}\left(0, 1 - \frac{(\Delta - 5)^2}{(30-5)^2}\right) & \text{if } \Delta > 5 \\ \text{MAX}\left(0, 1 - \frac{(\Delta + 5)^2}{(30-5)^2}\right) & \text{if } \Delta < -5 \\ 1 & \text{otherwise} \end{cases} .$$

As shown in Fig. 2, the score is between 0 and 1 and it would reach the maximum when the difference $\Delta = 60(\hat{t}_0 - t_0)/v$ is small, where the difference is proportional to the spindle speed. Note that we tend to overestimate $\Delta$ a little because of the difference in the surface speed between the internal and external tool-workpiece gap (due to the depth of cut).

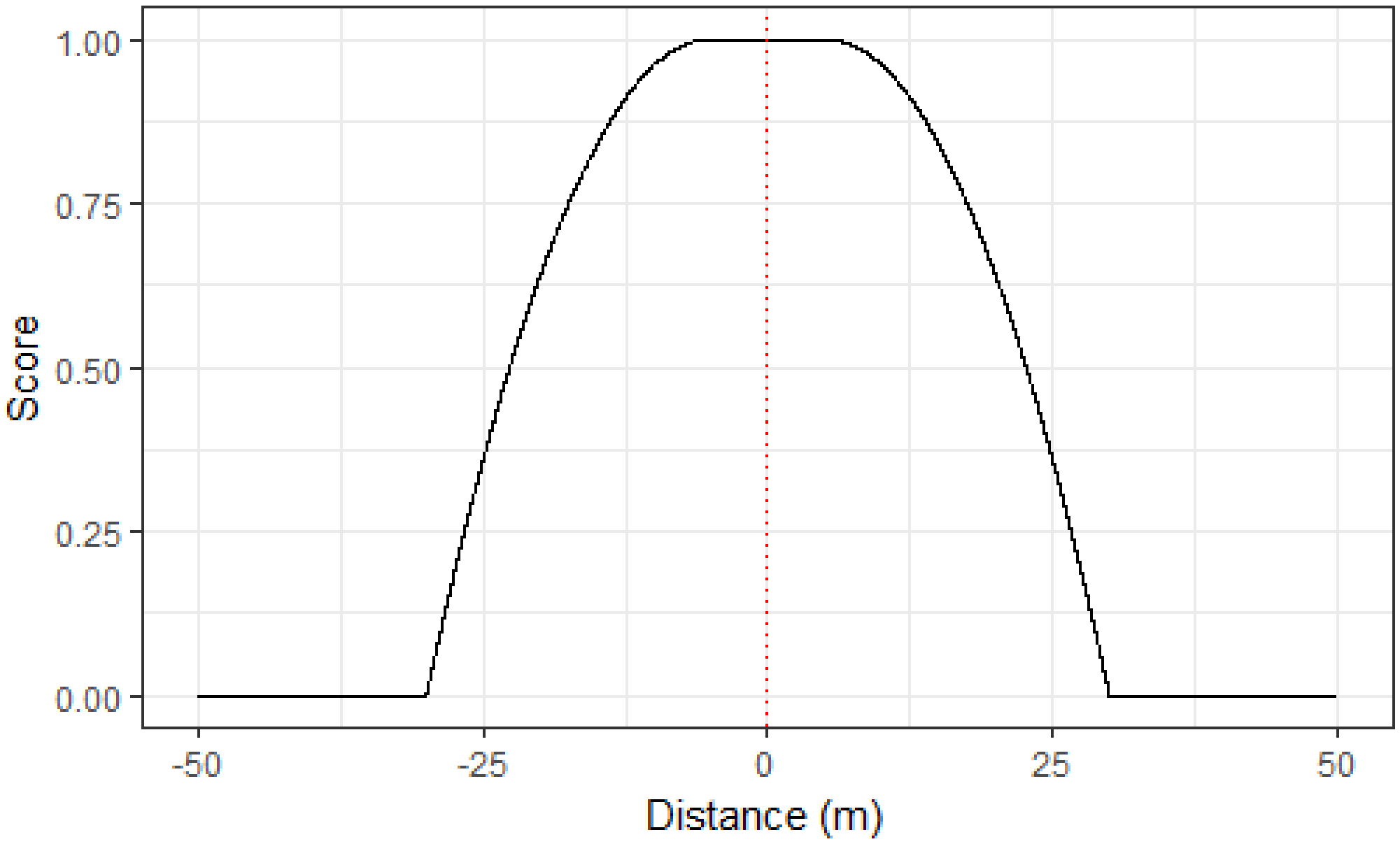

**Fig. 2** Our tool wear prediction score function. We allow a window of opportunity for the anomaly detection algorithm to identify an anomaly and score it according to the function above. An algorithm that predicts the distance as 0 will get the highest score of 1, whilst if the algorithm is more than 30 meters too early (in terms of cutting distance) then it will receive a score of 0, and same with being 30 meters too late

In this paper, our goal is to minimise $\Delta$ for the 12 out of the 16 progressive wear runs we select in random for training our real-time algorithm (e.g. to optimise its computer parameters and compare several anomaly detection methods), and thus maximise the average of the scores. We therefore apply the best scoring algorithm configurations to the remaining four test runs to evaluate its out-of-sample performance, as we demonstrate in Section 4.

### 2.2. Review of Anomaly Detection Methods

We look at different methods for detecting outliers and anomalous patterns, the potentials and weakness of which will be discussed here to select those that might work well in real-time tool wear monitoring. For more detailed and thorough review of mostly information-technology-related literatures, see Patcha and Park (2007), García-Teodoro et al. (2009), and Gupta et al.

(2014). Unfortunately, many conventional detectors (e.g. classification algorithms) are found to be infeasible for our tool wear analytics because they either do not work in a real-time environment (Ahmad et al. 2017), or they do not have easily implementable software interfaces. We broadly classify the commonly used anomaly detection methods into following categories:

1. Informal. Graphical and descriptive measures, e.g. from a data visualisation, data profiling, or exploratory data analysis.

2. Model-based. Most of these come from the statistical literature and use probabilistic mathematics to create e.g. predictive probability distributions for the current state of the data or processes. Methods here include both parametric and nonparametric modelling. The price to be paid for calculating full uncertainty distributions is often a more simplistic behaviour of the mean state of the data. Moreover, they understandably require massive computational effort in a real-time environment if the model must be validated on a continuous basis. This discourages their use in practice. The simplest model-based methods are control charts that are very commonly used for quality control.

3. Algorithm-based. The algorithm uses a set of logical reasoning steps to determine the current state of potential anomalies (and tools in our study). Such methods are usually chosen based on good practice on large multidimensional data, rather than probabilistic mathematics. Many of them tend to produce predictions without (or with simplistic) estimates of uncertainty. Examples also include unsupervised learning methods such as artificial neural networks and statistical data mining.

### 2.2.1. Statistical Process Control (SPC)

Lines between different categories of methods are sometimes blurred. For instance, process control (and statistical tests) is often considered as descriptive and less formal (like control charts) when in fact, its theoretical foundation is on the distribution assumption for a constant mean statistical model. Furthermore, if we build SPC into the outlier detector for tool wear monitoring, this evolves into an algorithm-based method suitable for real-time implementation.

To detect outliers and anomalies with SPC, one can use the classical $3\sigma$ rule (aka 68-95-99.7 rule) or one of its extensions, i.e. Western Electric rules, Nelson rules, or a custom set of our own rules under the $6\sigma$ principles for process improvements (Montgomery 2005; Montgomery and Woodall 2008), depending on the characteristics of the outliers we need to discern. There are R packages for control charts and visualisation, such as “qcc” (Scrucca 2004) and “qicharts2” (Anhøj 2018). Under the 68-95-99.7 rule, for instance, we should compute the control limits of the Shewhart control chart based on the moving ranges (MR) of all (or a portion of) historical force data, in which case we take their average $\overline{f}$ as the mean of the absolute first-order difference (FOD):

$$\overline{f} = \frac{\sum_{i=2}^{n} |F_i - F_{i-1}|}{n-1},$$

where $F_i$ indicates the force at the $i$th measurement and $n$ indicates the number of observations. The value of $\overline{f}$ is also an estimate of the scaled standard deviation, $1.128\,\mathrm{SD}(|F_i - F_{i-1}|)$, of MR. Static SPC makes sense if force values (or their MR) do not evolve over time and there are neither trends nor irregular oscillations. Without a model to describe such trends in force, we should focus on MR or FOD instead. After we include the extra factor 0.8525, as described in Harter (1960), the MR control limits around the central line are

$$\frac{(1.128 \pm 3 \times 0.8525)\overline{f}}{1.128}.$$

All we need for real-time application is to recalculate the average MR whenever we detect outliers, and there is no need to retain old data and measurements. We can revise the sigma level (3 for the above control limits) too, and this will be introduced in Section 3.1.

#### 2.2.2. Statistical Model-Based Methods

A model-based method relies on it being a continuous measure of the “outlier-ness” of each force measurement. Such methods include linear regression, time series models, nonparametric smoothing, and stochastic processes (Kelleher et al. 2015). For instance, one can fit a linear regression model and calculate the Cook’s distance of all data points, which is a measure of statistical leverage and deviation (Cook 1977). While this or other parametric models can describe linear and consistent data patterns, sometimes via using change point estimation (Wang and Samworth 2018), a model loses much of its practical sense when data patterns fast evolve over time and become too complicated to be modelled. A time series model such as autoregressive integrated moving average (ARIMA) might partially address this by trying to build outlier effects and irregular trends into the model (Chen and Liu 1993), but its main use is to forecast trends and uncertainties in the future (e.g. in financial markets or macroeconomics) so that it is recommended for tool wear monitoring. In contrast, nonparametric methods include splines, kernel smoothers, dynamical systems (Raftery et al. 2010), and Gaussian process regression (Bhinge et al. 2016). Besides, in practical engineering, one may do a frequency analysis, e.g. via Fast Fourier Transform, to determine the smoothing parameter values for signal filtering (Silverman 1982), which can be classified as a nonparametric smoothing noise-mitigating method. The idea behind most nonparametric methods is to offset high-frequency noise, so that the data are approximated by one or more smooth curves. For anomaly detection, this idea is however challenged by difficulties in justifying the nonparametric model (some hidden outlier effects might be removed because of information loss caused by smoothing the data), and in validating the model on real-time basis that would slow down computation. Hence, in Section 3, we rule out all the model-based methods for our demonstration.

#### 2.2.3. Statistical Probabilistic tests

Most statistical tests are faster than model-based methods and can be used to evaluate mean or variance deviation of data points. Grubbs’ test is a parametric outlier test for time series assumed to be from a normal or at least a lognormal distribution. Its two common variants are the Tietjen-Moore test (where the number of outliers must be fixed) and the generalised extreme studentised deviate (GESD) test (Rosner 1983), which allows us to find more than one outlier in time series. The GESD test is recommended when the number of outliers is unknown at a given time. It has, for instance, been used to monitor abnormal time-evolving trends and hot topics in Twitter (Vallis et al. 2014; Hochenbaum et al. 2017), though the Twitter algorithm also evaluates a seasonal effect for time series (non-exist in our tool wear monitoring). An alternative method is to use the Pearson's chi-squared (chi2) test (Ye and Chen 2001), which assumes a normal distribution and is similar to the Z-test of some SPC methods. Nonparametric outlier tests (Stephens 1974; Zhang 2002), such as the Kolmogorov-Smirnov test, Anderson–Darling test, and Cramér-von Mises test, focus on comparing the empirical distribution of two datasets (e.g. from two different time windows of the time series). There are also tests for comparing group variance, such as the Bartlett's test, Levene's test, and Cochran's C test. For brevity and simplicity, we will not further discuss these tests for group comparison, some of which are in the R package "outliers" (Komsta 2006). For our tool wear monitoring application, we will use the GESD test and chi2 test, which we consider as the most relevant and feasible real-time tests.

### 2.2.4. Dimension Reduction and Data Transformation

In the turning process we ran, three data streams of force were observed whereas other unmeasured variables such as temperature, spindle load, and vibration coexist. In practice, experimenters might monitor several factors and collect data corresponding to the same time windows. While individual time series can be studied one at a time, this is not feasible nor advisable in high dimensional cases when a multivariate approach should be used to take account of the correlation between time series (often for prediction or classification purposes). To facilitate detection and assessment of potential anomalies in high-dimensional cases, e.g. with tens of response variables of various nature, a wide range of relevant methods are discussed in Zimek et al. (2012) and Schubert et al. (2014), where dimension reduction such as the well-established principle component analysis (Hawkins 1974; Jackson and Mudholkar 1979; Aggarwal and Yu 2001) is useful.

When the dimension is low, and we have some prior knowledge about the data and the correlation between time series, dimension reduction can be as straightforward as merging individual time series into one or two data streams. For instance, the tangential force, vertical force, and feed force are all observed in our three-dimensional case, so a naïve dimension reduction solution would be to take the (weighted) average of the absolute of the three forces (i.e. linear combination), which can be used to calculate the Manhattan distance between forces at different times. In Section 5, we average the absolute of the tangential and feed forces, as we consider the vertical force to be uninformative and disturbing in our particular application. Linear combinations work best when the response variables (two force components in our case) are homogeneous with similar properties (e.g. consistent measurement method and unit). A similar solution is to use the Mahalanobis distance, the most well-known distance measure (Worden et al. 2000), though that would require extra work to compute the covariance matrix of data (based on an assumed normal distribution). This is undesirable for real-time monitoring.

Our real-time solution is to exclude the covariance matrix, in which case, the Mahalanobis distance between data points will revert to the Euclidean distance. The Euclidean distance is also regarded as an extension of MR/FOD to the multivariate case, and most important of all, in our application, it could be interpreted as a trace of the instantaneous variation in the resultant force (derived from the tangential and feed forces). We can then run SPC or statistical tests (i.e. GESD and chi2) to detect anomalies as in the univariate case. Furthermore, if we were to dismiss the timestamps (though we cannot in this real-time application), one might consider distance-based neighbourhood search or classification methods such as the k-nearest neighbour (Hautamaki et al. 2004), which are used to evaluate an outlier's distance to a number of its nearest neighbouring data points. Another alternative is to evaluate the densities (or spatial proximities) of data points instead of the distances. Local outlier factor is such a method that treats observations with lowest local densities as outliers (Breunig et al. 2000). However, these methods based on either distances or densities cannot be applied to data structure of time series and real-time data, when we should use SPC or statistical tests based on the Euclidean (Manhattan) distance.

The Euclidean distance is an extension of MR/FOD, but we also propose a new and more conservative distance measure here. For force observation $F_i$, a scalar in the univariate case (but extended to a 3-dimensional vector in the multivariate case), the first-order difference (aka backward difference) is calculated as $F_i - F_{i-1}$. An alternative is to consider both the forward and backward direction in time. To do so, we propose a new metric, the minimum successive difference (MSD), as $\mathrm{Min}_{\mathrm{abs}}\,(F_i - F_{i-1}, F_i - F_{i+1})$, where the minimum function is based on the absolute of the first-order backward and forward differences (i.e. to choose the shorter distance). If we aim to detect those lone outliers rather than a short sequence of successive outliers, MSD should be the better transformation. Either distance measure is more informative than the raw data because now we can test for outliers without too much doubt about original data patterns introduced through CNC machine programming and experimental procedure (often, it would require us to fit a complex model to interpret such patterns as discussed in Section 2.2.2). In the univariate case, we should monitor each data stream of FOD or MSD and look for the earliest anomalies (to predict tool wear and the best stopping time), whereas in the multivariate case, we focus on the combined data stream of the Euclidean distance.

### 2.2.5. Decision Trees, Neural Networks, and Deep Learning

Many of these methods are suitable for data of higher dimension, e.g. when there are more than five predictor variables in the dataset, though these do not apply here. For instance, decision tree methods (Stein et al. 2005) are often used for classification purpose and we can regard them as some variants of distance-based learning and clustering, allowing for extra flexibilities in arranging data structure and rules. Deep learning has potential for future manufacturing engineering (Chen and Lin 2014) but is currently inappropriate for our real-time application. Recurrent neural networks such as recent LSTM methods (Debar and Dorizzi 1992; Hawkins et al. 2002; Malhotra et al. 2015) are suitable to use often when the focus is on model prediction (rather than outlier recognition) and the data are far too complex (e.g. to investigate textual data from a foreign language) for conventional models and tests we have described in this section. Requiring plenty of time and data for training, classification and deep learning methods work best in a supervised setting, i.e. when we have detailed prior information about the classification mechanism. Comparatively, our study focuses on designing and applying a simpler set of necessary rules for anomaly detection in a small dataset.

## 3. Real-Time Algorithm and Decision Rules

None of the existing methods and software packages we find can be directly used without developing our own real-time solutions. Applying a generic anomaly detection algorithm is likely to lead to disappointing performance in manufacturing analytics. Therefore, we need to first build an efficient real-time algorithm that is capable of processing data streams (e.g. differencing and smoothing the raw data) and subsequently running data analytics. Depending on the timing and characteristics of the anomalies identified, the algorithm should evaluate the instantaneous tool wear level towards the end, consequently deciding on when a warning should be sent (ideally just before its estimated time of tool failure) to interrupt manufacturing. Our real-time algorithm should fulfil three key criteria:

1. It should be fast and robust in terms of integrating, processing, and managing a large volume of data over time. We must minimise computing time and, in a position where we examine anomalies (every second in our application), the detection should be simple enough without complex modelling steps or revisiting too much historical data.

2. It should take account of the natural time correlation and structure of data streams.

3. It should be flexible in the sense that we should be able to apply it elsewhere by tuning only a few meaningful parameters. It should also be generally compatible with anomaly detection rules that may be added into the framework and monitoring system.

The mechanism of our real-time detector should include three stages as illustrated in Fig. 3: 1) when a new batch of force data points arrive and read into the algorithm, data processing is required to clean the noise and transform raw force values to FOD (or MSD); 2) after an initial force threshold check, we assess the outlier-ness of all extreme values in FOD or MSD (advanced classification rules would be needed to detect anomalous time-evolving trends such as a gradual increase in force, which is not our focus in this outlier detection paper); 3) when one or more extreme force values are found as anomalies, we use them to real-time predict the tool wear level (and decide on the suitable timing to remove the worn-out or grossly fractured tool).

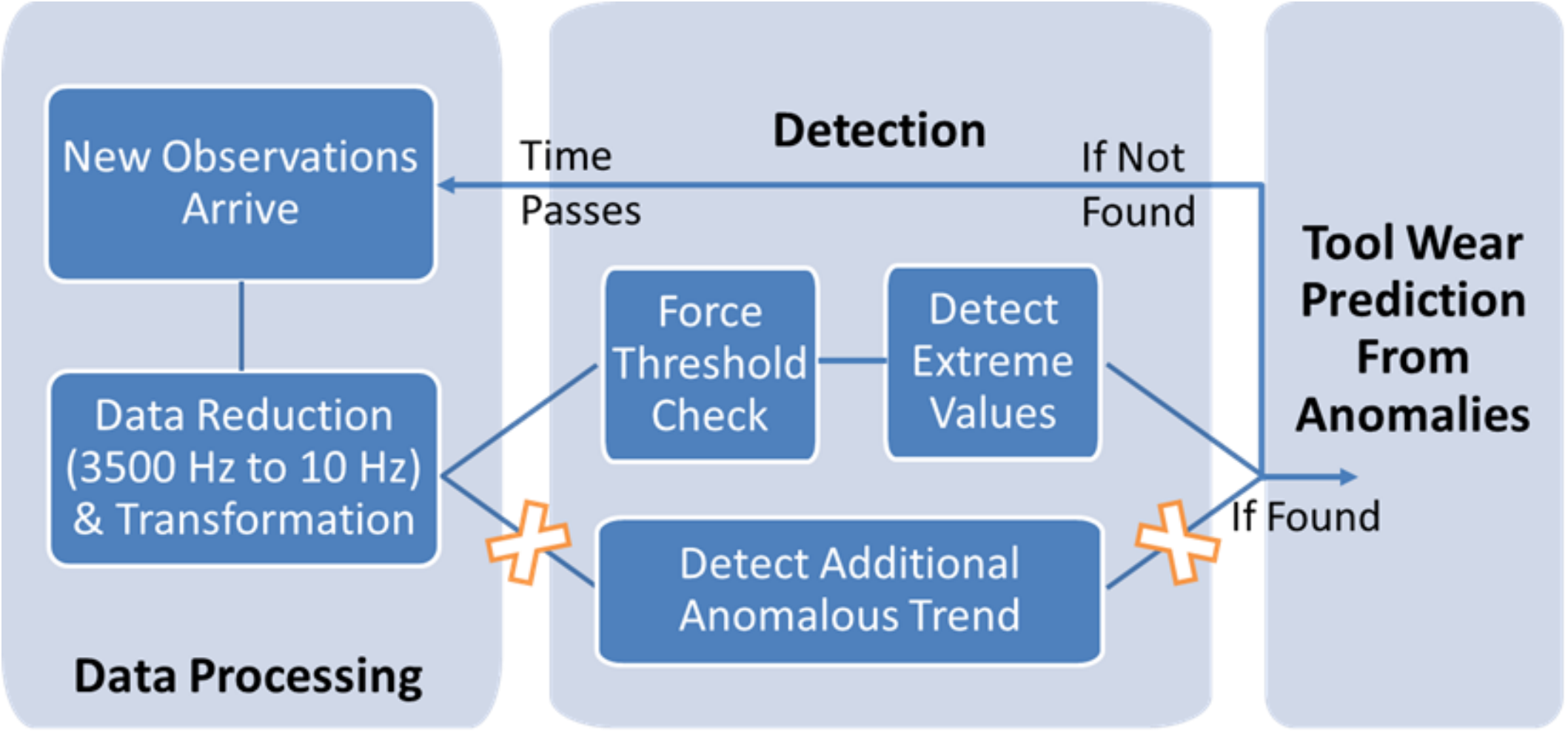

**Fig. 3** Structure of the Real-Time Tool Wear Monitoring Mechanism

As soon as a process starts, force measurements are made at 3500Hz, so that 3500 new observations should be added each second when the data stream is expanding over time. To facilitate data processing and better recognise the patterns, we reduce the size of raw data from 3500Hz to 10Hz through a “smart resampling” method. For each subset of 350 data points (collected in a duration of 0.1 second), we take their absolute values and convert them to one observation of the new 10Hz data, where we take the 90% quantile of the distribution of the subset (ensuring that a large but not extreme value is used as the representative of previous 350 observations). This is illustrated in Fig. 4 (a) and (b), as our goal is to convert data streams into some feature vectors for machine learning that are quite robust against potentially very rare outliers. Other than the 90% quantile we use, there are alternative metrics such as median (Basu and Meckesheimer 2007), mean, the 10% quantile, or the bandwidth (defined as the difference between the 90% and 10% quantiles in Fig. 4 (c)). All these metrics should be better than resampling the 3500Hz data at a fixed interval (aka systematic sampling), as this initial processing step removes high-frequency data patterns (as the variation is greatly reduced when the tool is in good health condition) and subsequently simplifies out task to locate anomalies or abrupt changes in the statistical profile (e.g. mean shift and variance increase) of the force data, before we proceed to calculate FOD or MSD.

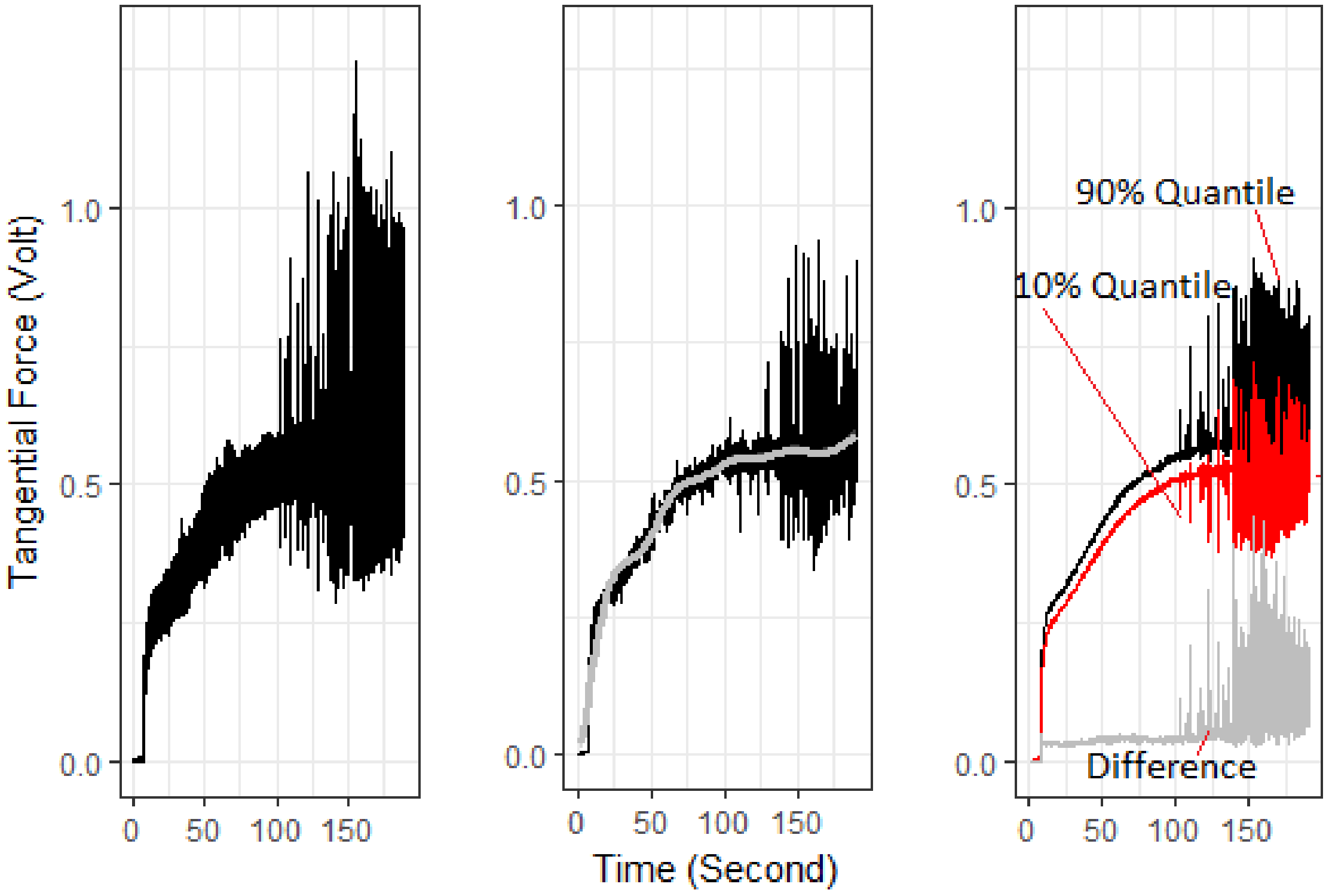

(a) (b) (c)

**Fig. 4** Measured tangential force in the 20th run of the turning experiment, which took less than 200 seconds (as indicated on the X-axis): (a) raw data stream at 3500Hz; (b) 10Hz data stream created via resampling the raw data stream at a fixed interval; (c) 10Hz data stream of the 90% quantiles (as well as the 10% quantiles just below the 90% trace and the pairwise differences or bandwidths near the bottom, which we include for reference)

In our off-line simulation with SPC, the dataset should be modified at the end of each second to include most recent observations (i.e. a 10-unit subset of the FOD (MSD) of the 90% quantiles). We then determine whether this subset as a whole is anomalous, which we define as a data vector $\mathrm{D}$. Let $\mathrm{Max}(|\mathrm{D}|, r)$ denote the $r$th largest absolute value of the 10 and $\mathrm{F}$ denote the corresponding 90% quantiles of force (in the multivariance case, the average of the tangential and feed forces). When there is little variation in force, anomalies should not exist, and the tool condition is considered to be stable. Therefore, in each second (or whenever a new batch of observations arrives), we start with an initial force magnitude check, where the thresholds are determined based on our intuitive choice and first training results. We evaluate $\mathrm{Max}(|\mathrm{D}|, 1) = \mathrm{Max}(|\mathrm{D}|)$, the maximum absolute value of FOD (MSD), and if it exceeds the threshold 0.05 volts, or if $\mathrm{Max}(|\mathrm{F}|) > 0.5$ volts, we should use one of the three methods below to detect the outlier-ness of the subset $\mathrm{D}$. Otherwise, we can move to the next time point and check the new 10 observations. Similar threshold-based detection methods are seen in industrial applications across all areas. In tool wear monitoring literature assuming and evaluating an empirical force model, thresholding (and sometimes differencing time series) ideas are seen in case studies such as Altintas and Yellowley (1989), Lee et al. (1995), Tansel et al. (1998), de Jesús et al. (2003), Shao et al. (2004), and Rehorn et al. (2005).

### 3.1. Real-Time Statistical Process Control

For SPC, we use a combination of four rules as follows and likewise, the values $\mathrm{Max}(|\mathbf{D}|, 2), \mathrm{Max}(|\mathbf{D}|, 4), \mathrm{Max}(|\mathbf{D}|, 6)$ corresponds to the second, the fourth, and the sixth largest value recorded in one second:

$$\begin{cases} \mathrm{Max}(|\mathbf{D}|,1) > \dfrac{(1.128 + 3\omega \times 0.8525)\mathrm{Mean}(|\mathbf{D}|_{\mathrm{acc}})}{1.128} \\ \mathrm{Max}(|\mathbf{D}|,2) > \dfrac{(1.128 + 2\omega \times 0.8525)\mathrm{Mean}(|\mathbf{D}|_{\mathrm{acc}})}{1.128} \\ \mathrm{Max}(|\mathbf{D}|,4) > \dfrac{(1.128 + 1\omega \times 0.8525)\mathrm{Mean}(|\mathbf{D}|_{\mathrm{acc}})'}{1.128} \\ \mathrm{Max}(|\mathbf{D}|,6) > \dfrac{(1.128 + 0\omega \times 0.8525)\mathrm{Mean}(|\mathbf{D}|_{\mathrm{acc}})}{1.128} \end{cases}$$

where $\omega$ is a weight parameter and the mean of the absolute of the accumulated historical observations in $\mathbf{D}$, $\overline{f} = \mathrm{Mean}(|\mathbf{D}|_{\mathrm{acc}})$, is the unbiased estimate of $1.128\ \mathrm{SD}(|\mathbf{D}|_{\mathrm{acc}})$. Since there is no need to calculate the standard deviation, we can delete the old data as time passes to boost computing and real-time updating. We choose a range of weights $\omega \in [0.8, 3]$ in the univariate case and half the standard weight in the multivariate case, because our distance metric is based on two force data streams, as mentioned in Section 2.2.4. Practitioners should be flexible and decide themselves which rules should be enabled for real-time SPC. For our demonstration in Section 4, we require all rules to be satisfied, in which case the small data subset (consisting of the 10 latest observations) can be treated as anomalous. When the first anomalous subset is found at time $\hat{t}_0 = t$, the tool wear level is estimated to be near 150 micrometres. This is a basic decision mechanism we envisage.

### 3.2. Chi-Squared Test

The chi2 test is similar to SPC, where we assume the distribution of $\mathbf{D}$ to be Gaussian. Shown in the set of rules below, $\mathrm{Max}(\mathbf{D}, 1)$ denotes the FOD (MSD) corresponding to the maximum absolute value thereof and this is allowed to take a negative value too:

$$\begin{cases} \dfrac{(\mathrm{Mean}(\mathbf{D}_{\mathrm{acc}}) - \mathrm{Max}(\mathbf{D}, 1))^2}{\mathrm{Var}(\mathbf{D}_{\mathrm{acc}})} > (3\omega)^2 \\ \dfrac{(\mathrm{Mean}(\mathbf{D}_{\mathrm{acc}}) - \mathrm{Max}(\mathbf{D}, 2))^2}{\mathrm{Var}(\mathbf{D}_{\mathrm{acc}})} > (2\omega)^2 \\ \dfrac{(\mathrm{Mean}(\mathbf{D}_{\mathrm{acc}}) - \mathrm{Max}(\mathbf{D}, 4))^2}{\mathrm{Var}(\mathbf{D}_{\mathrm{acc}})} > (1\omega)^2 \end{cases}.$$

Hence, the chi2 test in our demonstrations is based on the first rule of the above set, where the threshold is $9\omega^2$ and the weights $\omega \in [1.2, 2.6]$ are equivalent to those defined under SPC. Note that we must also evaluate the variance of the historical data in this case. If we use the RO scheme, which is more time-consuming than RW, a shortcut is to fast update the variance from the recursive relation between mean and variance.

### 3.3. GESD Test

We create a modified version of the GESD test, with three rules below:

$$\begin{cases} |\mathrm{Max}(\mathbf{D},1)-\mathrm{Mean}(\mathbf{D}_{\mathrm{acc}})| > \dfrac{n_1+1}{\sqrt{n_1+2}}\sqrt{\dfrac{\omega^2T^2}{n_1+\omega^2T^2}}\mathrm{SD}(\mathbf{D}_{\mathrm{acc}}) \\ |\mathrm{Max}(\mathbf{D},2)-\mathrm{Mean}(\mathbf{D}_{\mathrm{acc}})| > \dfrac{n_2+1}{\sqrt{n_2+2}}\sqrt{\dfrac{\omega^2T^2}{n_2+\omega^2T^2}}\mathrm{SD}(\mathbf{D}_{\mathrm{acc}}) \\ |\mathrm{Max}(\mathbf{D},4)-\mathrm{Mean}(\mathbf{D}_{\mathrm{acc}})| > \dfrac{n_3+1}{\sqrt{n_3+2}}\sqrt{\dfrac{\omega^2T^2}{n_3+\omega^2T^2}}\mathrm{SD}(\mathbf{D}_{\mathrm{acc}}) \end{cases},$$

where $T$ is the t statistic with $n_1 = n-2,\ n_2 = n-3,\ n_3 = n-5$ degrees of freedom corresponding to the different rules, and the respective significance are $(1-\mathrm{Prob}(3))/n, (1-\mathrm{Prob}(2))/n, (1-\mathrm{Prob}(1))/n$ (i.e. the probabilities associated with different quantiles of the t distribution). The significance is proportional to the number of observations fed to the GESD test, and it will approach zero as more and more observations are added to the data stream. As a result, RO does not work with the GESD test, as we seek to limit the maximum of $n$.

## 4. Results

We use 12 out of the 21 runs (the remaining four runs of progressive wear form a test set as indicated in Table 1) to tune the tool wear monitoring parameters (e.g. the weight) and decide the most promising configuration based on the tool wear prediction score function. In the univariate case, we look for the earliest anomalies found in the three force data streams. In the multivariate case, we calculate the Euclidean distance so that there is just one data stream for the resultant force. It is feasible to use a RO scheme to use all the historical data, we will also examine the usefulness of RW and in this case, RW is based on the last one minute's data (i.e. RW=60s). We compare the three detection methods (i.e. statistical process control, chi-squared test, and GESD test) on data of FOD and MSD. We then calculate the mean score based on the 12 runs and the tool wear prediction score function in Fig. 2, which is plotted in Fig. 5 against the weight. We also contrast the scores between rolling origin and rolling window, and between first order difference and mean successive difference.

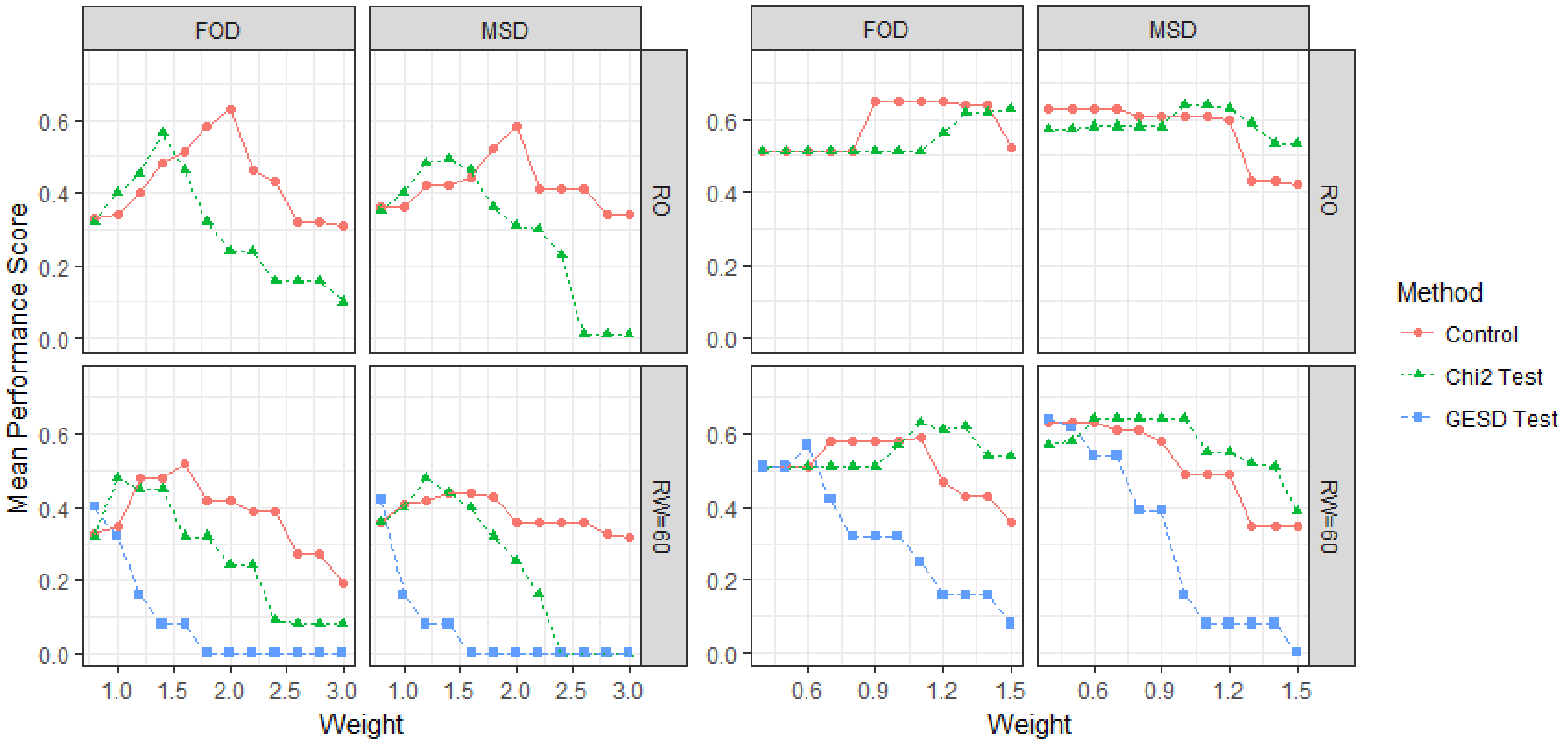

a) Univariate b) Multivariate

**Fig. 5** Mean performance score across the training data sets in the (a) univariate case where we monitor the time series of the three force components in parallel and stop at $\hat{t}_0$ when the first anomalous pattern is identified (b) multivariate case where the detection is based on the Euclidean distance (i.e. FOD or MSD of the tangential force and feed force)

We see from the scores that on average, SPC scores higher than the chi2 test in the univariate case when we check FOD (MSD) of each force component. In contrast, both methods work well in the multivariate case, as the mean score has been boosted to 0.65. Moreover, in the multivariate case where the Euclidean distance is calculated to approximate the FOD (MSD) of the resultant force, the algorithm performance seems to be more stable and less reliant on the weight and data transformation. The score remains high across different weights in particular if we use RO rather than RW (60 seconds per window). As also mentioned in Section 3.3, the GESD test cannot be run under the RO scheme, because its criterion would become too strict for outlier detection when the data starts to grow large. When we use RW, the GESD test is not as useful as the other two methods but it can also achieve reasonable performance if we choose a small weight in the multivariate case.

As we examine all the scenarios in Fig. 5, the highest mean score is just above 0.65 and the score is found to be robust to possible weight misspecification in the multivariate case. While these results are based on 12 out of the 16 progressive wear runs, we also evaluate a test score for the other 4 runs excluded. The mean test score for those best cases (where the mean score is near 0.65 in Fig. 5) varies between 0.55 and 0.60, and this indicates lower but reasonable performance when we consider the degree of uncertainties in real applications. Meanwhile, we can run a second test on the 5 runs of broken flank, though this should correspond to a different tool wear mode and we aim to predict the time when the tool is grossly fractured. The mean score is close to 1 under those best scenarios but it can reduce to 0.4-0.6 when we increase the weight to, for instance, 1.2 in the multivariate case.

Our recommendation is to calculate the Euclidean distance as in the multivariate case, where a smaller weight should be chosen if we use data of MSD instead of FOD. For an overview of individual runs, we take out a SPC FOD RO treatment at weight 1.2 and a chi2 test MSD RW treatment at weight 0.8. Predicted time $\hat{t}_0$ of all the 16 runs of progressive wear should be contrasted against the $\mathrm{VB} \approx 150\mu m$ time $t_0$ in Table 1, and the results are illustrated in Fig. 6. Note that both the tool wear time (and tool life difference) are approximations under the simplified assumption of linear progressive wear and cutting distance increase at uniform speed (and this approximation might be unreliable for the turning process so that the actual tool wear is unpredictable), the actual values of which are unknown and hard to measure as in practice.

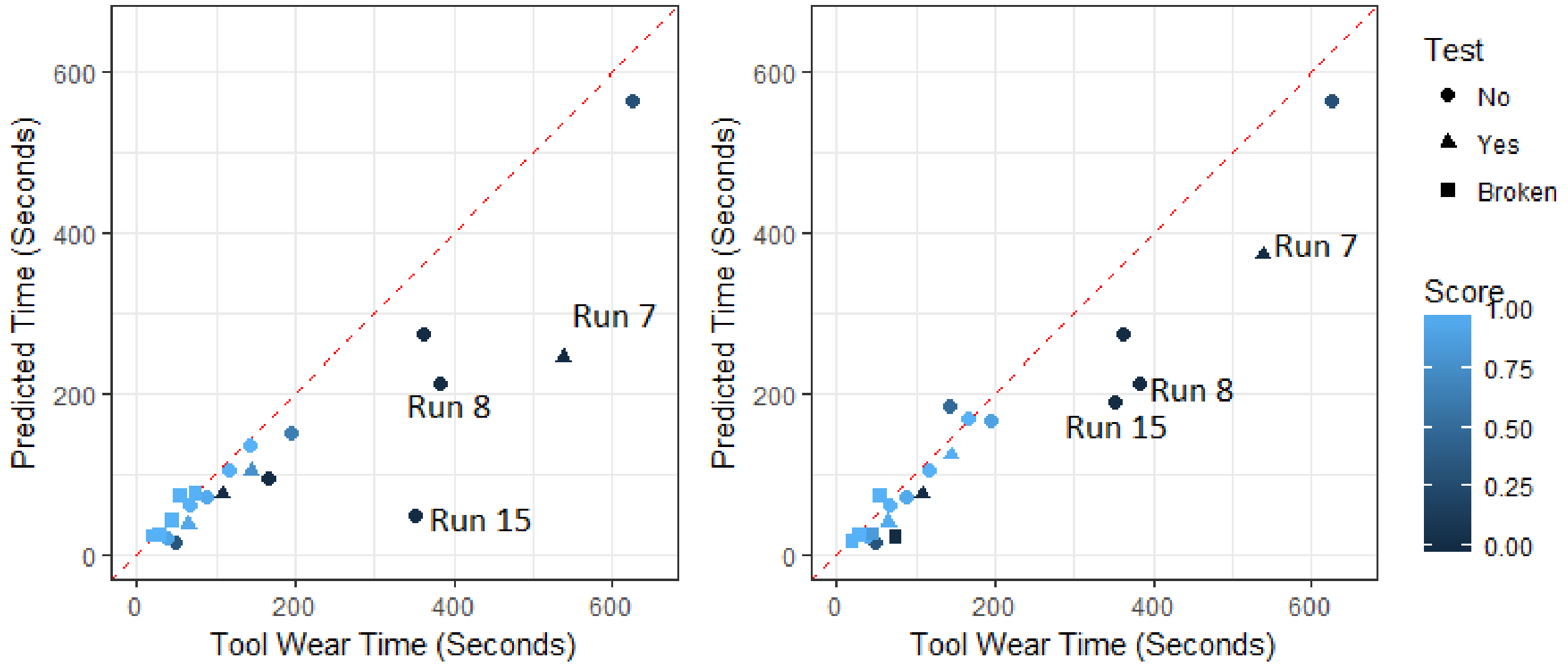

(a) SPC

(b) Chi2 test

**Fig. 6** Assumed tool wear time (the assumed $\mathrm{VB} \approx 150\mu\mathrm{m}$ time) versus predicted time. The face lightness of the points (the 5 triangles indicate the 5 test runs) reflects the score we calculate from the tool life difference

Most of the points/triangles in Fig. 6 are close to the dashed diagonal line and hence indicate high agreement between the assumed tool wear time and our prediction. However, there is a moderate gap in each of the runs 7, 8, and 15, when data indicates faster than expected tool wear increment. This could be due to either poor prediction from our real-time algorithm (i.e. we overestimated tool wear) or inaccurate assumption of strict linear progressive wear (i.e. tool wear reached 150 before the extrapolated time $t_0$). As we inspect runs 8 and 15, anomalies can be found at the beginning of a new pass requiring the workpiece to be reloaded in the cutter, which might be part of the problem that makes force variation more violent. In practice, we might need to disable real-time monitoring for these few seconds to avoid false warning of anomalies, if the whole process is expected to go on and last for more than a few minutes.

## 5. Discussion and Conclusions

While our real-time tool wear monitoring algorithm best performs in the multivariate case as demonstrated in Fig. 5, it seems to be robust across different scenarios too. We can see little difference in the calculated score between the two real-time streaming schemes (i.e. rolling origin and rolling window) although we expect computing to be faster under RW than RO in the case of some longer processes. We can also choose to calculate either the first-order difference of the force values or the minimum successive difference as proposed in this paper. The later is a more conservative measure of uncertainties and variation, but it leads to similar performance as we find in our demonstration in Section 4. Both statistical process control and chi2 test can work well in terms of detecting outlier forces and anomalous changes in force, allowing us to correlate such anomalies with high tool wear (and possible performance issues with the CNC machine). However, both methods have been modified to fit in tool wear monitoring, as we define our own rules and run the detection real-time as we describe in Section 3. The GESD test fails to produce comparable scores in this case, but this can be improved by adjusting its criterion and significance level (e.g. by further decreasing the weight parameter).

Overall, being capable of instantaneous learning from the concurrent force data streams, our real-time algorithm also has the artificial intelligence required to extract, transform, and examine the data patterns observed and associate them with tool wear, without interrupting the CNC-machining process. We believe it should contribute to any real-time tool wear monitoring system and ultimately help to automate the decision-making and on-line tool replacement process in manufacturing. More importantly, to facilitate real-time use and increase computing efficiency, our anomaly detection does not involve very complex mathematical calculations or predictive modelling, and the data updating process has been greatly streamlined. Based on the data streams of the cutting force, the algorithm can be used to identify the time when tool wear exceeds a certain threshold and further predict the time when tool failure or a suspected fault is expected to be imminent. As such, it is able to inform the computer control system to send timely warnings to machine operators.

As shown in Section 4, our prediction of the assumed tool wear time is generally reasonable and reliable (with a mean score above 0.65 and just three inaccurate predictions as indicated in Fig. 6), even though the actual tool life would be affected by both controlled variables (spindle speed and feed rate in our experiment) and varying environmental factors. Moreover, our anomaly detection methods might experience some performance issues in practice when:

1. There are no distinguishable anomalies from the time series data, and this might occur in the case of slow progressive wear and/or small measured force;
2. Tool health condition quantification (i.e. tool wear measurement) is inaccurate as this is often a different task in practice;
3. Our assumption of strict linear progressive wear is inappropriate because progressive tool wear without a "flank wear limit" may, for instance, exhibit an s-shape with a high initial rate followed by a near linear portion and finally accelerating wear towards the end of tool life.
4. Tool failure occurred too close to the start of the whole process (or the time when machine restarts with a new session), i.e. before sufficient force data can be harvested to monitor the ongoing tool wear;
5. Force data are collected at a low frequency or discontinuously, so that the algorithm suffers from information loss;
6. Anomalies exist but their features are too sophisticated too be identified using existing rules for real-time monitoring;
7. There are external issues, such as technical issues with the CNC machine or the electronic devises linked to the machine (e.g. for data acquisition purposes), change to uncontrolled environmental factors (e.g. temperature and pressure), and human errors, that affect the data we collect.

While little can be done to tackle items 1-5 above (perhaps except for using a reasonable tailor-made experimental and lab design), we should eliminate item 6 as much as possible and make item 7 under strict control for the manufacturing process. In practice, this requires using a large volume of data to calibrate and refine the real-time algorithm we propose here for new scenarios (e.g. outside the current range of indicated fundamental process parameters) and "boundary conditions" of the experiment. As more data displaying different properties and structures are used to stepwise optimise anomaly detection, we would have better understanding of any association and causality between tool wear and the measured force, aligning with the on-line quality control principles for continuous improvement. We may also learn by observing the varying condition of the scrapped tools (e.g. to better quantify tool health condition and possibly tell how to prevent tool failure) and redesigning the process for future experiments.

To detect different tool wear mechanisms and modes, one can also add new outlier classification rules, for instance, to evaluate anomalous trends and time-evolving patterns. This might demand additional pattern recognition work to model the data off-line (and hope some insights of practical or economic value can be transferred to on-line data analytics and real-time tool wear prediction). Note that this cannot be done on real-time basis because model re-evaluation is computationally very expensive generally, especially in the case of multi-dimensional deep learning, and may require additional data. This draws our attention for further research in the near future.

## Acknowledgements

We would like to show our gratitude to the Machine Tool Technologies Research Foundation (MTTRF), DMG Mori and DP Technology for the award of the excellent facilities used for this research.

We would like to thank DePuy Synthes and Enterprise Ireland for supporting this research through the Innovation Partnership (IP) programme (no. IP/2016/0428). The Innovation Partnership programme is co-funded by the European Union through the European Regional Development Fund 2014-2020.

Andrew Parnell is also part of the SFI I-Form centre for Advanced Manufacturing funded by Science Foundation Ireland and the European Regional Development Fund.